%% file: main.tex
\documentclass[conference]{IEEEtran}
\IEEEoverridecommandlockouts

\usepackage{cite}
\usepackage{amsmath,amssymb,amsfonts}
\usepackage{algorithmic}
\usepackage{textcomp}
\usepackage{threeparttable}
\usepackage{booktabs}

\usepackage{hyperref}
\usepackage{pifont}
\newcommand{\cmark}{\ding{51}}
\newcommand{\xmark}{\ding{56}}
\input{supplementary}

\def\BibTeX{{\rm B\kern-.05em{\sc i\kern-.025em b}\kern-.08em
    T\kern-.1667em\lower.7ex\hbox{E}\kern-.125emX}}

\begin{document}

\title{Blockchain and Artificial Intelligence:\\Synergies and Conflicts
}

\makeatletter
\newcommand{\linebreakand}{%
  \end{@IEEEauthorhalign}
  \hfill\mbox{}\par
  \mbox{}\hfill\begin{@IEEEauthorhalign}
}
\makeatother

\author{
    \IEEEauthorblockN{Leon Witt}
    \IEEEauthorblockA{\textit{Dept. of Comp. Sci. \& Tech.} \\
    \textit{Tsinghua University \& Fraunhofer HHI}\\
    % Beijing, China \\
    leonmaximilianwitt@gmail.com}
    \and
    \IEEEauthorblockN{Armando Teles Fortes}
    \IEEEauthorblockA{\textit{Dept. of Comp. Sci. \& Tech.} \\
    \textit{Tsinghua University}\\
    % Beijing, China \\
    fmq22@mails.tsinghua.edu.cn}
    \linebreakand
    \IEEEauthorblockN{Kentaroh Toyoda, \textit{Member}, \textit{IEEE}}
    \IEEEauthorblockA{\textit{Inst. of High Performance Computing} \\
    \textit{A*STAR} \\
    % Singapore, Republic of Singapore \\
    kentaroh.toyoda@ieee.org}
    \and
    \IEEEauthorblockN{Wojciech Samek, \textit{Member}, \textit{IEEE}}
    \IEEEauthorblockA{\textit{Dept. of Elec. Eng. \& Comp. Sci.} \\
    \textit{TU Berlin \& Fraunhofer HHI}\\
    % Berlin, Germany \\
    wojciech.samek@hhi.fraunhofer.de}
    \and
    \IEEEauthorblockN{Dan Li, \textit{Member}, \textit{IEEE}}
    \IEEEauthorblockA{\textit{Dept. of Comp. Sci. \& Tech.} \\
    \textit{Tsinghua University}\\
    % Beijing, China \\
    tolidan@tsinghua.edu.cn}
}

\maketitle

\begin{abstract}
Blockchain technology and Artificial Intelligence (AI) have emerged as transformative forces in their respective domains. This paper explores synergies and challenges between these two technologies. Our research analyses the biggest projects combining blockchain and AI, based on market capitalization, and derives a novel framework to categorize contemporary and future use cases. Despite the theoretical compatibility, current real-world applications combining blockchain and AI remain in their infancy. 
\end{abstract}


\section{Introduction}

In recent years, blockchain technology and Artificial Intelligence (AI) have independently driven transformative developments: blockchain technology has redefined the notion of trust through a decentralized, transparent, and immutable ledger technology~\cite{nakamoto2008bitcoin, buterin_ethereum_2014}, while the advent of deep learning~\cite{lecun_deep_2015} and scaling of large models~\cite{brown_language_2020} in AI has enabled many practical applications that enhance human capabilities, e.g., natural language processing~\cite{vaswani_attention_2017, devlin_bert_2019}, image recognition~\cite{krizhevsky_imagenet_2012, he_deep_2016},  autonomous systems~\cite{bojarski_end_2016}, and generation~\cite{goodfellow_generative_2014, kingma_auto-encoding_2014, ho_denoising_2020,openai2024gpt4, META2023llama}. On a conceptual level, blockchain technology appears to complement AI's limitations. Specifically, blockchain's inherent decentralization contrasts with AI's centralization issues; its transparent and verifiable nature addresses the opacity of AI models, and its robust data management capabilities support AI's data dependency. A recent rise in market capitalization of \textit{Blockchain X AI} cryptocurrencies surged to billions of USD, as shown in Fig.~\ref{fig:market_cap_crypto}, highlighting its positive market sentiment and growing investor confidence. However, practical integration reveals conflicts like the computational and storage overhead that contrasts with the distributed ledger architecture of blockchain, where every node stores and computes the same information redundantly.
This work analyzes technical synergies and conflicts between blockchain and AI to cluster the plethora of novel \textit{Blockchain X AI} use cases, as they vary in terms of use cases, system design, resolved issues, level of blockchain and AI integration, and availability of information. Instead of analyzing academic papers, where most projects remain theoretical \cite{BxAI_2019_IEEE, LeonSP, SP_Blockchain_IEEE, SP_MD_IEEE}, this work takes a bottom-up approach by focusing on existing \textit{Blockchain X AI} projects with a market capitalization exceeding 10 million USD as well as projects with unique use cases. The remainder of this paper is organized as follows: Section~\ref{sec:synergies_and_conflicts} details the synergies and conflicts between blockchain and AI technologies. Section~\ref{sec:use-cases} categorizes the state-of-the-art use cases at the intersection of blockchain and AI. Section~\ref{sec:discussion} discusses the limitations and future directions, with Section~\ref{sec:conclusion} concluding the paper.

\begin{figure}[t]
\centering
\includegraphics[width=\linewidth]{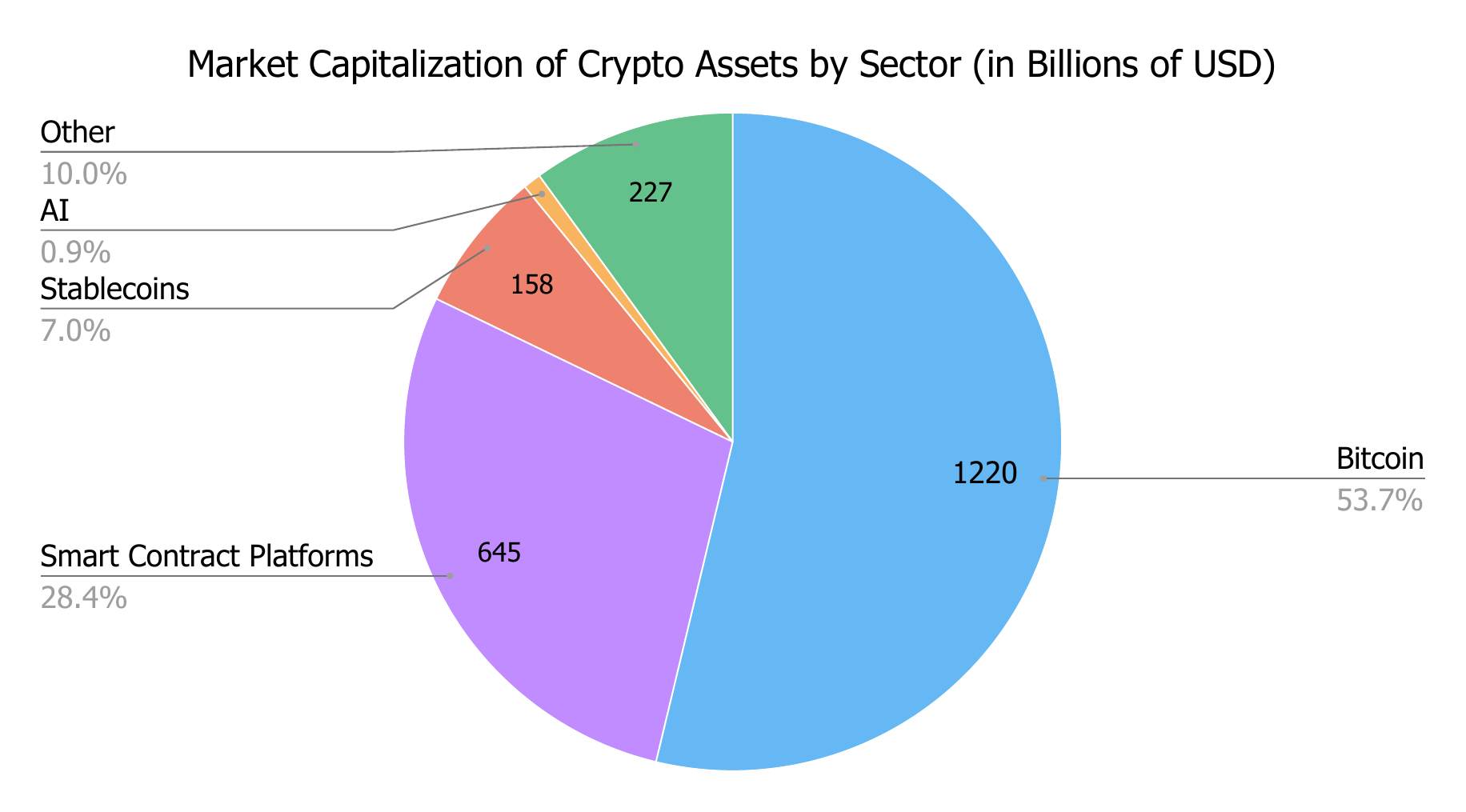}
\caption{\textbf{Market Capitalization of Crypto Assets by Sector.} The AI sector has a market capitalization of approximately 20 billion US Dollars (USD), accounting for 0.9\% of the total crypto market, which is valued at 2.27 trillion USD, representing an emergent niche.\protect\footnotemark}
\label{fig:market_cap_crypto}
\end{figure}
\footnotetext{Source: \url{https://coinmarketcap.com/}, accessed on April 18th, 2024.}

\section{Blockchain $\times$ AI: Synergies and Conflicts}
\label{sec:synergies_and_conflicts}

This section examines the synergies and conflicts between blockchain and AI, with Fig.~\ref{fig:BxAI_conflicts_synergies} illustrating their complementary and opposing characteristics.

\begin{figure}[t]
\includegraphics[width=\linewidth]{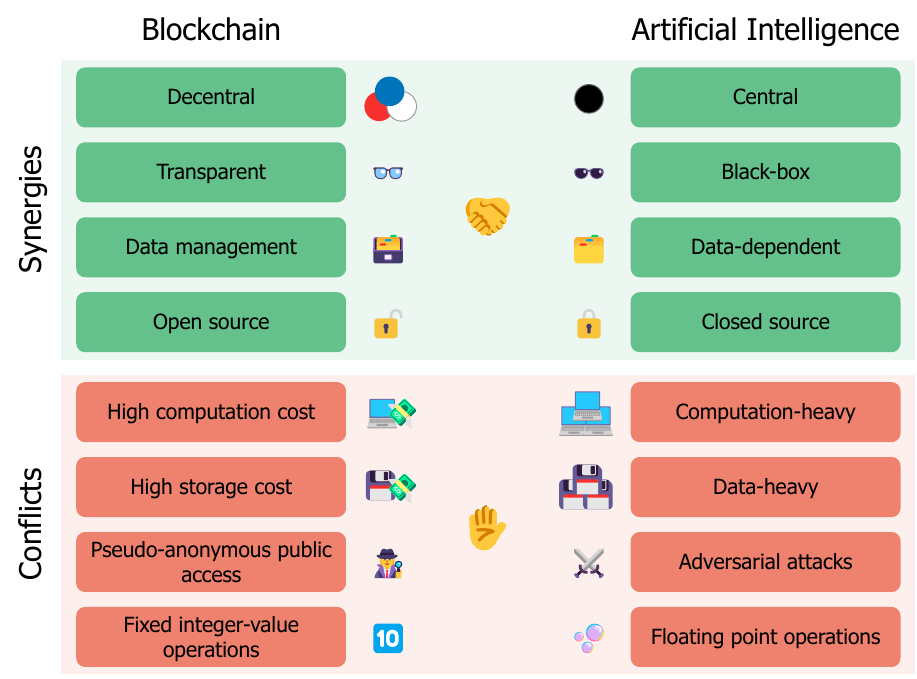}
\caption{Synergies and Conflicts between Blockchain and AI. }
\label{fig:BxAI_conflicts_synergies}
\end{figure}

\subsection{Synergies Between Blockchain and AI}

\textbf{Decentralization vs. Centralization.} State-of-the-art large language models like GPT require substantial computational and electrical resources as well as vast amounts of data to train and maintain. The computational cost for training GPT-3 in 2020 was approximately 4.6 million USD~\cite{Li2020GPT3}. This high cost has led to a concentration of capability within a few big-tech companies, effectively monopolizing the AI market. Such monopolization can hinder economic competition, a significant concern for policymakers in regions like the United States and the European Union, where antitrust laws are actively enforced to maintain market balance and prevent market dominance by a single entity~\cite{Sykes2022Antitrust, Hoerber2022Routledge}. In contrast, blockchain technology is inherently decentralized, presenting a potential counterbalance to the centralization typical in AI systems. The decentralized nature of blockchain ensures that no single party can control the network, thereby democratizing the data and operations managed on the blockchain. This feature could be particularly useful as a regulatory mechanism within AI by promoting a more balanced distribution of power and facilitating a more collaborative environment. Incorporating blockchain could, therefore, address ongoing public and regulatory debates concerning the dominance of major entities in AI, making the governance of such technologies more inclusive and equitable.

\textbf{Transparency vs. Black-box Nature.} Blockchain technology is synonymous with transparency, offering an immutable and verifiable record of transactions. This transparency could revolutionize the `black-box' nature of AI, wherein the reasoning behind decisions often remains obscured~\cite{samek2017explainable}. Blockchain's ledger can act as a platform for documenting AI decision-making processes, creating a transparent audit trail that enhances AI applications' trustworthiness. Additionally, it can incorporate advanced encryption techniques such as Zero-Knowledge proofs like Zero-Knowledge Succinct Non-Interactive Argument of Knowledge (zk-SNARKs)~\cite{Pinto2020zkSNARKs} or utilize secure hardware like Trusted Execution Environments (TEE)~\cite{TXX}. These technologies help verify that specific computational steps—such as the use of a particular AI model for inference—have been carried out honestly and accurately.

\textbf{Data Management and Dependence.}
Blockchain can regulate data and its access through smart contracts and protocols like the InterPlanetary File System (IPFS)~\cite{benet2014ipfs}, where data gets unique identifiers, ensuring data integrity and availability for AI systems to leverage. This synergy ensures that AI has access to high-quality, tamper-proof data, which is critical for the development of robust AI models. 

\textbf{Open Source vs. Closed Source.} By facilitating shared ownership through cryptographic protocols that enable fine-grained privacy configurations, blockchain could help address the limitations of AI's proprietary models. If successful in offering competitive business models, these shared AI systems, collectively trained and controlled by participants, could enhance transparency in AI development. This approach could encourage a broader range of contributions, potentially leading to the creation of more unbiased and comprehensive AI solutions.

\subsection{Conflicts Between Blockchain and AI}
However, despite these synergies, significant conflicts between blockchain and AI's operational requirements pose challenges to their integration.

\textbf{Computational Costs and Heavy Loads.} Training AI models and even performing inference operations on  Large Language Models (LLMs) like GPT-4 \cite{openai2024gpt4} or Llama 3\cite{META2023llama}, necessitates substantial computational resources. Blockchain's consensus mechanism, cryptographic operations, and unfavorable data structures add computational burden, negatively affecting scalability.

\textbf{Storage Constraints and Data Intensiveness.} The decentralized nature of blockchain, while ensuring security and redundancy, leads to increased storage requirements that can prove costly and inefficient for data-driven AI systems. In general-purpose blockchain systems (GBPS) such as Ethereum~\cite{wood2014ethereum}, every node in the chain has to store all information, as redundancy is part of the security and resilience of the blockchain network, which hinders scalability (Section~\ref{BlockchainAsInfra}). As new data has to be stored in a transaction format, common data based on the Ethereum virtual machine (EVM) structures like Patricia-Merkle-trees~\cite{wood2014ethereum} might impede optimal retrieval speeds. AI applications, on the other hand, generate and process large volumes of data, requiring efficient and scalable storage solutions.

\textbf{Pseudo-anonymity and Security Challenges.} Blockchain allows for permissionless, pseudo-anonymous access applying asymmetric encryption, with the network protected from Sybil attacks~\cite{SybillAttack} through computational~\cite{nakamoto2008bitcoin} or financial~\cite{wood2014ethereum} barriers that guard against adversarial behavior. In use cases where Blockchain serves as a platform to enhance privacy-preserving and decentralized AI training~\cite{LeonTSC} via methods such as Federated Learning (FL)~\cite{BrendanMcMahan2017}, allowing clients to participate in the training process pseudo-anonymously may pose risks. These methods are vulnerable to adversarial FL attacks~\cite{bhagoji2019analyzing}, where identifying malicious actors is particularly challenging because contributions to the overall AI model are private by design and not easily measurable~\cite{LeonSP}.

\textbf{Operational Mismatch.} Most blockchain virtual machines are designed with fixed integer operations to ensure deterministic outcomes, which is crucial when billions of USD are at stake in financial transactions or contractual agreements. This is because floating-point operations can lead to precision loss during calculations, particularly when numbers of vastly different magnitudes interact \cite{floatingpoint}. However, a common practice in AI training involves normalizing model floating-point parameters between 0 and 1, as it promotes stable and effective gradient flow and provides implicit regularization, thereby enhancing overall training~\cite{batchnorm}.

\section{Blockchain $\times$ AI: Use Case Study}
\label{sec:use-cases}

\begin{figure*}[t]
    \centering
    \includegraphics[width=\textwidth]{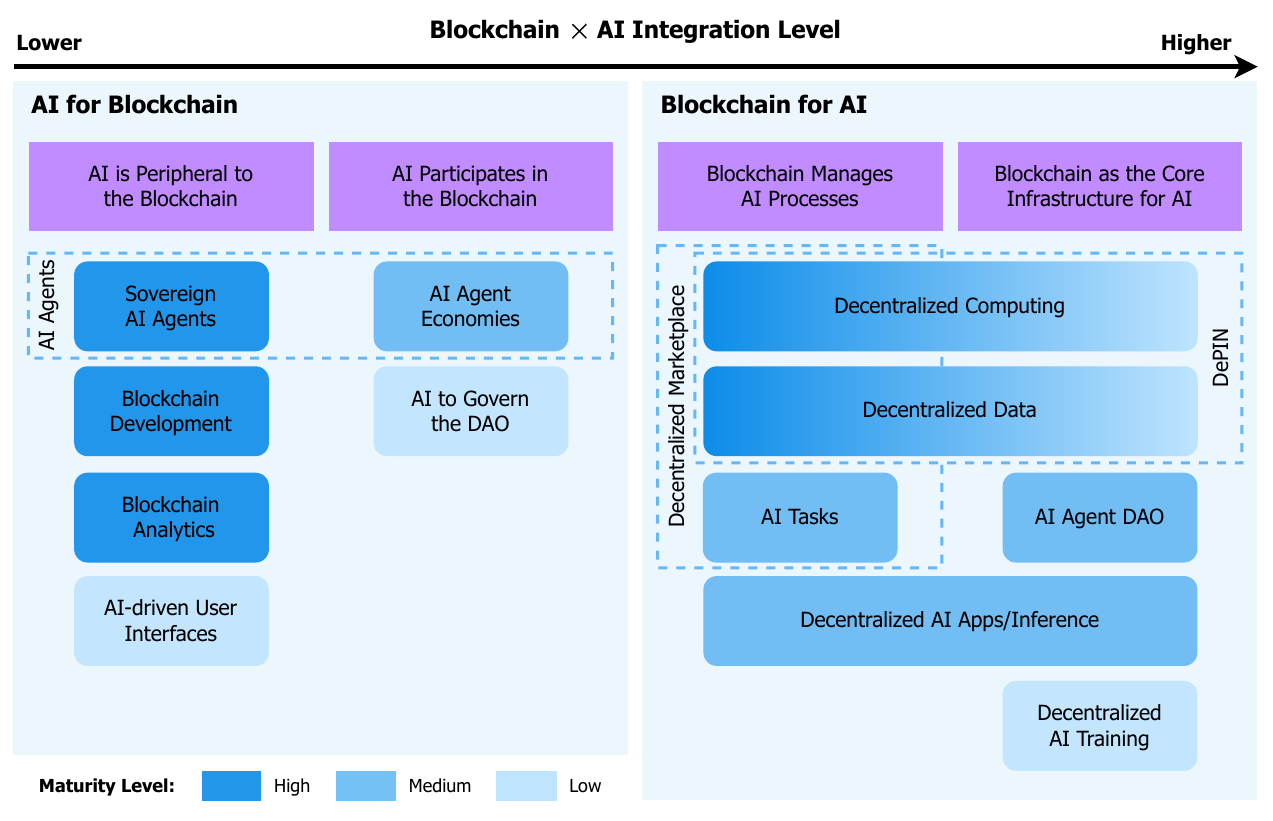}
    \caption{\textbf{Categorization of Use Cases at the Intersection of Blockchain and AI.} Two main categories are distinguished: `AI for Blockchain,' where AI supports blockchain functionalities, and `Blockchain for AI,' where blockchain technologies enhance AI processes and infrastructure. The horizontal axis shows the growing degree of blockchain integration — from AI as a peripheral tool to blockchain as the core infrastructure for AI.}
    \label{fig:BxAI_usecasecategories}
\end{figure*}

Given the aforementioned synergies and conflicts between blockchain and AI, the challenge of identifying projects that extend beyond theoretical concepts to demonstrate real-world use cases persists~\cite{BxAI_challenges, LeonSP}. We study state-of-the-art projects that integrate blockchain and AI. They vary significantly in terms of use cases, system design, resolved issues, level of blockchain and AI integration, and availability of information. This complexity opens a multitude of perspectives from which these projects can be analyzed. To the best of our knowledge, there is no consensus within the scientific community on how to categorize and analyze said projects. The majority of existing literature, mainly academic papers, tends to focus on theoretical aspects and is not production-ready yet~\cite{BxAI_2019_IEEE, LeonSP, SP_Blockchain_IEEE, SP_MD_IEEE}. This paper adopts a bottom-up approach, focusing on \textit{Blockchain X AI} projects that possess public tokens with a market capitalization exceeding 10 million USD\footnote{Source: \url{https://coinmarketcap.com/}, accessed on April 15th, 2024.}. Moreover, we also include noteworthy projects with market capitalizations below 10 million USD if they target novel use cases uncommon among existing projects.
We categorize use cases by addressing the following research questions aimed at the selected projects:
\begin{enumerate}
    \item What is the level of synergistic integration between blockchain and AI technologies within the project?
    \item What role does blockchain play within the project?
    \item What role does AI play within the project?
\end{enumerate}
Our analysis has identified four major clusters, as depicted in Fig.~\ref{fig:BxAI_usecasecategories}, representing use cases where:
\begin{enumerate}
    \item AI is peripheral to the blockchain;
    \item AI participates in the blockchain;
    \item Blockchain manages AI processes;
    \item Blockchain is the core infrastructure for AI.
\end{enumerate}

\subsection{AI is Peripheral to the Blockchain}
AI can play a supportive peripheral role even if it has no direct function within the Blockchain system itself. Peripheral use cases range from enhancing the user experience when interacting with the blockchain with AI, enabling analytics, and using AI support to streamlining blockchain development.

\textbf{Sovereign AI Agents.} Public blockchain systems and their underlying smart contract functionalities have introduced Decentralized Finance (DeFi) and prediction markets. Sovereign AI agents are lightweight software agents that individuals can run locally or in the cloud, performing specific tasks on the blockchain. The emergence of decentralized exchanges (DEXs) like UniSwap~\cite{uniswap2024}, coupled with the ease of introducing novel tokens (such as fungible ERC20 tokens on Ethereum), has created significant arbitrage opportunities within the DeFi sector. These opportunities, valued in the billions, have enabled AI agents to automate arbitrage strategies across various DeFi protocols~\cite{DeFiLlamaTVL2024}. Another promising use case is prediction markets, which are exchange-traded markets where individuals can bet on the outcome of various future events, such as presidential elections or sports scores~\cite{Polymarket2024}. Blockchain infrastructure supports decentralized prediction markets through transparent rules enforced by smart contracts and underlying cryptocurrencies, ensuring trust through transparency and immutability, as no third party can manipulate the outcome. A future use case could train AI agents to make predictions incentivized through monetary rewards and improved through Reinforced Learning with crowdsourced feedback~\cite{Buterin2024CryptoAI}. 

\textbf{Blockchain Development.} The advancement of Large Language Models (LLMs) has led to significant productivity boosts in coding and software engineering. However, expert domain knowledge remains essential, particularly in novel technologies at the cutting edge of cryptography research, such as Zero-Knowledge (ZK) Proofs, Homomorphic Encryption (HE), and Multi-party Computation (MPC). Despite their innovative nature, open-source projects are susceptible to vulnerabilities and errors. AI-coding like GitHub Copilot~\cite{GitHubCopilot2024} can play a pivotal role in identifying vulnerabilities, optimizing smart contract logic, and maintaining code for both smart contract developers and core infrastructure, as well as streamlining blockchain development by offering real-time suggestions, refactoring code, and automating repetitive tasks.

\textbf{Blockchain Analytics.} The integration of AI into blockchain analytics has opened new avenues for extracting insights and enhancing decision-making. AI algorithms can process vast amounts of blockchain data, facilitating on-chain analytics such as tracking transaction volumes, identifying trends in token transfers, and monitoring wallet activity~\cite{TheGraph2024}. Additionally, AI can contribute to market sentiment analysis by scraping social media platforms, forums, and news outlets to gauge public opinion on various blockchain projects and cryptocurrencies~\cite{cryptosentiment,cryptosentiment2}. 

\textbf{AI-driven User Interfaces.} The absence of a central entity to rectify mistakes when interacting with the blockchain, combined with its complexity, makes interactions with decentralized applications prone to human error and vulnerable to malicious actors and scammers. AI can enhance the user experience by simplifying complex concepts into understandable language, providing real-time guidance, and preventing user errors~\cite{Buterin2024CryptoAI}. For instance, AI-driven interfaces can assist users by explaining blockchain transactions in plain language, guiding them through decentralized application (DApp) interactions, and highlighting potential risks, which reduces the likelihood of errors or malicious activity. Additionally, AI can help detect suspicious transactions, thereby mitigating fraud.

\subsection{AI Participates in the Blockchain}

Building on the concept of AI as a supportive tool, this section explores the more integrated roles of AI within the blockchain framework, where AI actively participates in the blockchain's ecosystem and governance structures.

\textbf{AI Agent Economies.}
In this model, AI agents are not simply tools but rather participants or stakeholders within a decentralized network, where they act autonomously and interact or collaborate under the governance of collective community protocols like DAOs, exemplified by Fetch AI\cite{fetchai2024} and Olas~\cite{autonolas2024}.

\textbf{AI to Govern the DAO.} Another use case is the integration of AI into the governance of Decentralized Autonomous Organizations (DAOs), serving as an arbiter for decision-making or even making direct judgments~\cite{Buterin2024CryptoAI}. This can range from determining the acceptability of work submissions to deciding the correct interpretation of a constitution. However, this application requires careful consideration since AI's role in governing DAOs introduces several challenges, particularly concerning adversarial machine learning~\cite{dong2018boosting, szegedy2014intriguing, goodfellow2015explaining}. If the AI model is closed-source, its inner workings remain opaque, effectively functioning like a centralized application. Conversely, if the model is open-source, it becomes susceptible to targeted attacks, where adversaries can simulate and optimize attacks locally before replaying them on the live network. A potential solution to mitigate these risks involves the use of encryption schemes, such as Trusted Execution Environments~\cite{TXX} or zero-knowledge proofs~\cite{Pinto2020zkSNARKs}.

\subsection{Blockchain Manages AI Processes}

Blockchain technology is increasingly used to manage AI processes, creating a decentralized framework for resource sharing, data management, and application deployment.

\textbf{Decentralized Computing.} In the context of decentralized computing, blockchain can facilitate the sharing of processing power, such as GPUs, across a distributed network. Participants can rent out their idle computing resources to others, creating a peer-to-peer network that enhances accessibility to computational power. This is beneficial for researchers, data scientists, and small organizations that may not have the financial means to invest in expensive hardware or cloud services. Participants list their available computing resources on a blockchain platform, and users requiring these resources can secure them through smart contracts. These contracts automatically manage the terms of service, usage, and payments, ensuring a seamless transaction without the need for intermediaries. Blockchain projects leveraging this model include Akash Network~\cite{akash2024}, IO.net~\cite{ionet2024}, NodeAI~\cite{nodesai2024}, GamerHash~\cite{gamerhash2024}, Golem~\cite{golem2024}, Blendr Network~\cite{blendr2024}, Nosana~\cite{nosana2024}, CUDOS~\cite{cudos2024}, and DeepBrain Chain~\cite{dbcwiki2024}.

\textbf{Decentralized Data.} Similarly to computing resources, blockchain enables the creation of decentralized marketplaces for data. In this case, data providers can securely share their data, maintaining control over its usage while enabling access for purposes such as training AI models. Interested parties can purchase or access the data through smart contracts, which enforce the agreed-upon conditions, handle payments, and ensure compliance with data governance standards. Ocean Protocol~\cite{oceanprotocol2024} develops decentralized data exchange protocols to buy and sell data using data NFTs and data tokens, enabling token-gated access control, data DAOs, and data wallets. LayerAI~\cite{layerai2024} and Masa~\cite{masa2024} utilize ZK-rollups to ensure secure data storage and sharing. Synesis One~\cite{synesisone2024} aims to establish a Web3 data utility and marketplace governed by a DAO, focusing on crowd-sourcing data for AI systems. An alternative approach is presented by Grass~\cite{grass2024data}, which decentralizes bandwidth by allowing individuals to monetize their unused internet connection, facilitating large-scale web data scraping for AI model training.

\textbf{AI Tasks.} Analogously to previous use cases, businesses and individuals can outsource AI-related tasks (e.g., data collection and labeling) to a distributed network of contributors in decentralized marketplaces. This approach enhances the scalability of task completion and ensures diversity in data handling, which is critical for training unbiased AI models. PublicAI~\cite{publicai2024} operates a distributed AI network that connects businesses with a global pool of workers for a variety of AI tasks, from simple data annotation to more complex research-oriented tasks. Sapien AI~\cite{sapien2024}, on the other hand, integrates data labeling with gaming, creating an engaging environment where users are rewarded within a web3-based game.

\textbf{AI DApps.} DApps run autonomously, typically hosted on a blockchain and operating through smart contracts in an autonomous, always available censor-free way. Recently, AI-powered DApps have emerged, predominantly focusing on replicating proprietary AI models in a decentralized format. This approach is especially evident in generative AI (e.g., LLMs and diffusion models), enabling global and uncensored access~\cite{corcel2024, imgnai2024, monai2024}. Additionally, an intriguing example of AI integration in DApps is Botto~\cite{botto2024}, a decentralized autonomous artist that creates art based on collective community feedback via a DAO.

\subsection{Blockchain as the Core Infrastructure for AI}
\label{BlockchainAsInfra}

\begin{table*}[t]
    \centering
    \caption{Blockchain as the Core Infrastructure for AI. \textit{\footnotesize (DAI = Decentralized AI, BC = Blockchain, DT/FL = Decentralized Training/Federated Learning, C-Layer = Computational Layer, TA = Technical Analysis, DM = Distributed Management, PoS = Proof of Stake, DPoS = Delegated Proof of Stake, dBFT = Delegated Byzantine Fault Tolerance, FL = Federated Learning, DID = Decentralized Identity, ZK = Zero Knowledge, DePIN = Decentralized Physical Infrastructure Networks, DC = Distributed Computing, DD = Distributed Data, ASBS = Application Specific Blockchain System, IPFS = InterPlanetary File System, * = low maturity/no public code, ? = unavailable information).}}
    \setlength\extrarowheight{3pt}
    \resizebox{\textwidth}{!}{
    \label{table:BxAI_infrastructure}
    \rowcolors{2}{tableoddrow}{tableevenrow}
    \begin{tabular}{c|ccccccccc}
        %\toprule
        \rowcolor{tableheader} \textbf{Ref.} & \textbf{BC Type} & \textbf{Use Case} & \textbf{BC Function} & \textbf{Storage} & \textbf{Consensus} & \textbf{Token} & \textbf{DT/FL} & \textbf{C-Layer} & \textbf{TA} \\
        
        Bittensor~\cite{bittensor}& ASBS & DAI Training & Payment, DM & Validation Scores & Yuma
        consensus\cite{bittensor} & AI models, Validation, Governance & \xmark & \cmark & \xmark \\

        ORA~\cite{ORA2024} & opML (L2)  & Decentralized Inference & DM, (Payment) & Inference outputs & PoS (Ethereum) & \xmark (native Ether) & \xmark & \cmark & \xmark \\
        
        %Metadata (Merkle-Root, Fraud-Proofs)
        Axonum~\cite{Axonum} & opML (L2) & Decentralized Inference & DM, (Payment) & Inference outputs & PoS (Ethereum) & \xmark (native Ether) & \xmark & \cmark & \xmark \\

        Flock~\cite{Flock2024} & ASBS & DAI Training & Governance, Payment &  Metadata, Validation Scores & stake-based majority vote\cite{FLockIOPaper} & AI models, Validation, Governance & \cmark (FL) & \cmark & \cmark \\ 
        
        BalanceDAO~\cite{BalanceDAO2024} & ASBS & AI Agent DAO & Payment, Governance & Metadata, Reputation & PoS & AI models & \xmark & \cmark & \xmark \\

        DeepBrain Chain~\cite{DeepBrainChainWhitepaper} & ASBS & DC & DM, Governance, Payment & & DPoS & AI jobs/Inference & \xmark & \cmark & \cmark \\
        
        BasedAI~\cite{basedAI} & ASBS & ZK-Inference & Payment, DM, Validation & Metadata, ZK-Proofs &  TFT Enforcers\cite{basedAI} & Encryption, Inference & \xmark & \cmark & \xmark \\
        
        Devolved*~\cite{devolvedai2024whitepaper} & ASBS & DAI Training, DID & Payment, ? & ? & Proof of Value, PoS & ? & \cmark (FL) & \cmark & ?   \\
        
        Zero1 Labs*~\cite{z1labs2024litepaper} & ASBS & AI Agent DAO & Payment & ? & PoS & data, AI models & \xmark & \cmark & \xmark \\
        
        AIOZ~\cite{AIOZ2024} & ASBS & DePIN (DD,DC,AI Models) & Payment, DM & Pointer (IPFS) & dBFT & AI models, datasets & \cmark(FL) & \cmark & \xmark \\
        
        GPT Protocol~\cite{GPTProtocol2024} & ASBS & DAI, DD & Payment, DM & ? & PoS & AI models, Inference & \cmark & \cmark & \xmark \\
        
        Neura~\cite{neura2024}& ASBS & DC, AI Agent DAO & Payment, DM & Pointer (IPFS, URLs off-chain), Proofs & PoS (CometBFT\cite{CometBFTConsensus2024}) & GPU provision, AI models & \xmark & \cmark & \xmark \\
        
        Delysium*~\cite{delysium2023aiagents} & ASBS & AI Agent DAO & DM & Metadata & ? & AI models, governance & \xmark & \cmark & \xmark \\
        
        0G Labs~\cite{zeroGai}& L2 & Fast Data Layer & DM?& ? & ? & \xmark & \xmark & \xmark & ($\times $50k faster*) \\
        
        SingularityNET~\cite{singularitynet2019} & ASBS & DAI & DM, Payment & Metadata,  Pointer (IPFS) & Proof of Reputation & AI models, Inference, validation & \cmark & \cmark & \xmark \\
        
        Fluence~\cite{fluence2024}& ASBS & DePIN (DC) & DM, Proof of Capacity & Metadata, Proofs &  Proof of Compute & ? & \xmark & \cmark & \xmark \\        
        
        Orai Chain~\cite{oraichain2024} & ASBS & AI Oracles (AI Agent DAO) & DM, Security & Metadata(Oracle Scripts)& PoS & Governance, AI training & \xmark & \cmark & \xmark \\
        
        Vanna Labs~\cite{vannalabs2024} & ASBS & Decentralized Inference & Verification, DM& Proofs & ? &  Inference & \xmark & \cmark & \xmark \\
        
        Phoenix~\cite{phoenixglobal2024} & ASBS & DC &DM, Payment, Security& Metadata & \xmark & AI jobs/Inference & \xmark & \cmark & \xmark \\
        
        Galadriel~\cite{galadriel2024} & ASBS & Decentralized Inference & DM of AI-Agents &  Metadata (AI-Oracle Calls) & PoS (Tendermint) & \xmark (?) & \xmark & \cmark & \xmark \\
    \end{tabular}
    }
\end{table*}

General-purpose blockchain Systems (GPBS),  such as Ethereum~\cite{wood2014ethereum}, the largest public GPBS, face trade-offs between scalability, security, and decentralization~\cite{nakai2023blockchain}. Its consensus mechanisms are secured by thousands of active nodes around the world~\cite{EtherscanNodeTracker2024}, and almost a million staked validators~\cite{BeaconChainValidators2024}. As each new block of information must (i) reach and (ii) be verified by every node in the global network to attain consensus and finality, the average block size is in the kilobytes~\cite{EtherscanBlockSizeChart2024} and is created every 12 seconds which leads to high storage and computation costs. Hence, it is not practical to perform or store computationally intensive AI operations directly on-chain; however, a paradigm shift towards Layer-2 rollups, which process transactions off-chain and post aggregated results back, offer a cost-effective solution by increasing throughput and reducing costs~\cite{Arbitrum,optimism2024,scroll,zksync}. Similarly, blockchains dedicated to AI use cases must either (i) overcome challenges related to high computation and storage costs, public access, and the restrictions of the underlying virtual machine (Fig.~\ref{fig:BxAI_conflicts_synergies}) or (ii) use the blockchain as a mere managing, governing, and security layer. Table~\ref{table:BxAI_infrastructure} examines novel systems where blockchain is used as its core infrastructure.

\subsubsection{Blockchain Type} One solution is the Application-Specific Blockchain System (ASBS) for AI that functions as a Layer-1 solution; Layer-1 in blockchain denotes the primary network infrastructure responsible for processing and validating transactions through a designated consensus mechanism. In addition to ASBS, two compelling paradigms that can integrate into existing blockchains like Ethereum have emerged: zero-knowledge machine learning (zkML) and optimistic machine learning (opML), akin to the so-called optimistic \cite{Arbitrum, optimism2024} and zk roll-up systems \cite{scroll,zksync}. In opML, AI inference is abstracted out to a deterministic ML Engine compatible with the EVM, where every proposed result is considered valid by default. To protect against invalid results, the system includes a challenge period during which participants can contest the result and submit fraud proofs if conflicts arise \cite{conway2024opml}. Current solutions \cite{Axonum, ORA2024} store results on-chain in byte encoding. On the other hand, zkML is based on zero-knowledge proofs, particularly zk-SNARKs \cite{zk_proof}, which verify computations, such as ensuring the use of a specific, unbiased model for inference without exposing the data or AI model details \cite{Buterin2024CryptoAI}. Despite their promises, both approaches face drawbacks: While opML is fast, storing results on-chain is costly, and the finality of the results is delayed, as accepting any result as correct by default—being 'optimistic'—comes at the cost of potential errors later on (challenge period). While zkML offers instant finality due to its mathematically unambiguous cryptographic proofs, it also enables broader applications because only the zk-proofs are stored on-chain. Although these proofs can be created off-chain, they still incur substantial computational overheads. For instance, AI-critical operations like multiplication incur a 4× overhead, and non-linear operations such as the ReLU function can face up to a 200× overhead \cite{zk_proof, Buterin2024CryptoAI}.

\begin{table}[t!]
\caption{Comparison between opML and zkML~\cite{conway2024opml}.}
\label{tab:opMLvszkML}
\centering
\setlength\extrarowheight{2pt}
\rowcolors{2}{tableoddrow}{tableevenrow}
\resizebox{.8\linewidth}{!}{
\begin{tabular}{c|cc}
\rowcolor{tableheader} & \textbf{opML} & \textbf{zkML} \\
\textbf{Model Size} & Unlimited & Limited \\
\textbf{Computation} & Low & High \\
\textbf{Storage} & High & Low \\
\textbf{Finality} & Delayed & Instant \\
\textbf{Security} & Medium & High \\
\textbf{Mechanism} & Crypto-economic  & Cryptographic  \\
\textbf{Proof} & Fraud-proof & Validity-proof \\
\end{tabular}
}
\end{table}

% Despite their promises, both approaches face drawbacks: While opML is fast, storing results on-chain is costly, and the finality of the results is delayed, as accepting any result as correct by default—being 'optimistic'—comes at the cost of potential errors later on (challenge period). While zkML offers instant finality due to its mathematically unambiguous cryptographic proofs, it also enables broader applications because only the zk-proofs are stored on-chain. Although these proofs can be created off-chain, they still incur substantial computational overheads. For instance, AI-critical operations like multiplication incur a 4× overhead, and non-linear operations such as the ReLU function can face up to a 200× overhead \cite{zk_proof, Buterin2024CryptoAI}.

\subsubsection{Use cases} To run AI training in a decentralized fashion (DAI)~\cite{FLockIOPaper, devolvedai2024whitepaper, singularitynet2019, AIOZ2024}, through methods like FL, new blockchain architectures are required. Decentralized Inference~\cite{bittensor, oraichain2024, vannalabs2024, galadriel2024}, where users pay to use pre-trained AI models for inference—often in conjunction with encryption techniques like Trusted Execution Environments~\cite{galadriel2024} or zero-knowledge proofs~\cite{basedAI} to ensure data integrity and correct execution of the respective AI model—requires less complex ASBS as it manages existing AI-nodes, if the model does not reside on the blockchain but on dedicated layers. Similarly, ASBS that manage and govern AI Agents is a major use case~\cite{BalanceDAO2024, z1labs2024litepaper, delysium2023aiagents, neura2024, oraichain2024}. Blockchains optimized for Decentralized Physical Infrastructure Networks (DePINs)~\cite{fluence2024, z1labs2024litepaper} include Distributed Computing (DC)~\cite{dbcwiki2024, phoenixglobal2024, neura2024, fluence2024}, where clients provide GPUs for AI training, and Distributed Data (DD)~\cite{GPTProtocol2024, AIOZ2024}, where edge nodes contribute their data to the training.

\subsubsection{Blockchain functionality}
The major functions of the ASBS for AI are payments, Distributed Management (DM) or governance of DAO-like structures, validating zero-knowledge proofs, and verifying certain operations performed by clients while acting as an immutable and transparent log.

\subsubsection{Storage} Most projects store application-specific metadata to ensure an immutable trace of governance and actions performed within the system on the blockchain. As storage is one of the major scalability bottlenecks due to its redundant nature, large amounts of data like model parameters are usually stored off-chain through IPFS pointers that link to the respective data lanes~\cite{AIOZ2024, neura2024}. Furthermore, either validation scores—where validators are rated based on AI model inference outputs \cite{bittensor} or the quality of training \cite{Flock2024}—or proofs, such as zk-SNARKs produced off-chain, are stored \cite{basedAI, vannalabs2024}.

\subsubsection{Consensus} Next to common consensus mechanisms like Proof of Stake (PoS), where validators vouch with their money on the correctness of new information being added to the blockchain, novel approaches have been invented to better fit the respective needs of AI. Flock~\cite{Flock2024}, which aims to decentralize FL, lets randomly selected validators verify the current model. The update is appended to the blockchain if the majority of the clients agree. Bittensor's~\cite{bittensor} Yuma Consensus rewards subnet validators with dividends for producing miner-value evaluations that are in agreement with the subjective evaluations produced by other subnet validators, weighted by stake. Temporal Fusion Transformers (TFT)~\cite{basedAI} enforce rules and rewards based on contributions over multiple timesteps, similar to the Proof of Reputation of SingularityNet~\cite{singularitynet2019}. DevolvedAI~\cite{devolvedai2024whitepaper} seeks to reward clients by creating value within the system (Proof of Value), though specifics remain unclear. Fluence's~\cite{fluence2024} Proof of Compute enforces the generation of cryptographic proofs for all executions in the network. Providers generate proofs for their computation, and customers pay only for the work accompanied by proofs that the work was validated and correct.

\subsubsection{Decentralized Training, Federated Learning, and Computational Layer} Federated Learning~\cite{BrendanMcMahan2017} appears to be the predominant method for training a global AI model $\theta$ in a decentralized manner, where a central server aggregates locally trained models $\theta_i$ according to:
\begin{equation}
    \arg\min_{\boldsymbol{\theta}} \sum_{i} \frac{|S_i|}{|S|} f_{i}(\boldsymbol{\theta}).
\end{equation}
Here, for each client $i$, $f_{i}$ represents the loss function, $S_i$ denotes the dataset of each client, and $S:=\bigcup_{i} S_i$ represents the combined set of indexes of data points of all participants. Blockchain technology can replace the central server to further decentralize the FL process. Witt et al.~\cite{LeonSP} categorize the functions of blockchain as a model aggregator, coordinator, payment, and storage entity, based on a systematic literature review of decentralized and incentivized FL frameworks. To run AI training in such a decentralized fashion requires a redesign of contemporary GPBS, as the high computational demands and storage costs would make it infeasible to be fully deployed on blockchains where every node has to compute and store every model update. ASBS that want to decentralize AI training should include (i) restrictive access, (ii) an adapted consensus mechanism, (iii) the separation of computation from the blockchain, and (iv) a data layer with scalable storage and retrieval technologies to manage the heavy computations and storage demands of modern deep neural networks. 

\subsubsection{Token} One of the major advantages of a blockchain system is its inherent cryptocurrency functionality, which makes it easy to automate payments through smart contracts, incentivizing actions like AI model provision~\cite{bittensor, BalanceDAO2024, AIOZ2024, GPTProtocol2024, delysium2023aiagents, singularitynet2019} to validate and verify the quality of participating models~\cite{bittensor,singularitynet2019}, conduct inference on large language models~\cite{dbcwiki2024, basedAI, GPTProtocol2024, singularitynet2019, vannalabs2024, Axonum, phoenixglobal2024} on provided data, and encourage the creation of zero-knowledge proofs~\cite{vannalabs2024} to ensure correct execution of AI models.

\subsubsection{Technical Analysis and Experiments} Decentralization inevitably causes overhead in terms of speed and complexity. Only a few of the projects have peer-reviewed publications on theoretical foundations while none have published experiments or tests about their performance.

\section{Results, Limitations \& Future Outlook}
\label{sec:discussion}

Despite the several billion USD in market capitalization, most \textit{Blockchain X AI} projects are still in early development. The following outlines challenges and considerations:
\begin{enumerate}
    \item \textbf{Maturity of Use Cases.} Emerging use cases such as AI-agent economies, AI governance for DAOs, and ASBS for AI demonstrate potential but are early in their maturity. The full realization of these applications and whether they emerge as compelling use cases remain uncertain.
    \item \textbf{System Complexity.} Generally speaking, the deeper the blockchain integration, the more complex the systems, particularly ASBS for AI, which necessitates domain knowledge for set-up and operation.  (Table~\ref{table:BxAI_infrastructure}). With a multitude of competing projects in the blockchain and AI domain, identifying and differentiating viable projects becomes increasingly difficult.
    \item \textbf{Adaptation Needs.} The advent of advanced generative AI technologies like GPT-4~\cite{openai2024gpt4} and accessible, open-source LLMs such as LLaMA~\cite{META2023llama}, which may surpass privately trained models, questions the viability of decentralized marketplaces for private models. Existing \textit{Blockchain X AI} projects may need to adjust their use cases to better align with current technological advancements.
    \item \textbf{Technical Difficulties.} Technical challenges remain in achieving privacy-preserving, decentralized, and democratized AI (Fig.~\ref{fig:BxAI_conflicts_synergies}). While the ASBS is trying to overcome said conflicts by introducing novel blockchain architectures with adjusted consensus, data, and computation layers, they come at the cost of complexity and computational overhead.    
    \item \textbf{Regulatory Uncertainty.} Regulatory uncertainties remain, such as whether decentralized AI training and Federated Learning comply with strict data privacy laws in the US and EU.
\end{enumerate}

\subsection{Future Outlook}

Applications of existing blockchain systems, where AI becomes a participant, enabling precise and micro-level operations, not only appear promising and straightforward to implement but also open up a wide array of future applications through use cases where autonomous AI agents can be controlled and rewarded by various DAOs. Conversely, the higher the level of integration between blockchain and AI, the more complex the technical challenges become, and hence, the less mature the technology is. Achieving a single democratic governance for an overarching, systematically-important AI, upon which other applications depend, involves using deep blockchain integration as well as cryptographic methods. While these approaches have the potential to enhance functionality and AI safety without centralization risks, they also face significant uncertainties regarding their foundational assumptions. It remains to be seen whether these systems will scale beyond niche use cases, where security is paramount, to achieve mass adoption. The required domain knowledge and complexity could pose significant entry barriers, and it is uncertain whether consumers will tolerate performance sacrifices compared to centralized solutions due to blockchain overhead.

\section{Conclusion}
\label{sec:conclusion}

The integration of blockchain technology and AI represents a promising yet complex frontier. Although there is considerable potential for synergy, and despite significant recent developments in some of the conditioning variables, practical applications often encounter conflicts due to differing system designs. Given the wide variation in use case, complexity, blockchain integration, system design, maturity, and thoroughness among existing projects, this paper presents a categorization framework and discusses the challenges that must be addressed to progress from a minimum viable product to mass adoption.

\section*{Acknowledgment}

We would like to express our gratitude to Pengyun Che and Ricardo Teles Fortes for their insightful discussions on emergent use cases for \textit{Blockchain X AI}, which greatly enhanced the depth of this research. We thank Rosanna Bassett for proofreading the manuscript.

\bibliographystyle{IEEEtran}
\bibliography{bibliography}

\end{document}

%% file: supplementary.tex
\usepackage[table]{xcolor}
\usepackage{graphicx}

\usepackage{standalone}
\usepackage{amsmath,amssymb,amsfonts}
\usepackage{algorithmic}
\usepackage{textcomp}
\usepackage{babel}
\usepackage{setspace}
\usepackage{balance}

\usepackage{varwidth}
\usepackage{verbatim}

\usepackage{tikz}
\usetikzlibrary{positioning,shapes.geometric,fit,calc}
\usepackage{pgfplots}
\pgfplotsset{compat=1.16, every non boxed x axis/.append style={x axis line style=-},
     every non boxed y axis/.append style={y axis line style=-}}
     
\usepackage[edges]{forest}
\usepackage{caption}

\tikzset{parent/.style={align=center,text width=3cm,rounded corners=3pt},
    child/.style={align=center,text width=3cm,rounded corners=3pt}
}

\colorlet{col1}{white}
\colorlet{col2}{gray!15}
\colorlet{col3}{gray!30}
\colorlet{col4}{gray!40}
\colorlet{col5}{gray!50}

\colorlet{transparent}{white!0}
\definecolor{ieeeblue}{RGB}{0,120,177}
\definecolor{accessblue}{cmyk}{1 0.3 0 0.2}
\definecolor{accessDarkBlue}{HTML}{006699}
\definecolor{tabletopheader}{gray}{.6}
\definecolor{tableoddrow}{gray}{.95}
\definecolor{tableevenrow}{gray}{.9}

\definecolor{tableheader}{HTML}{65b8f2} % light blue
\definecolor{tableoddrow}{gray}{.975}
\definecolor{tableevenrow}{gray}{.95}